\documentclass[letterpaper, 10 pt, journal, twoside]{IEEEtran}

\usepackage{graphicx}
\usepackage{amsmath}
\usepackage{array}
\usepackage{multirow}
\usepackage{hyperref}
\usepackage{booktabs}

\begin{document}

\title{PIE: Parkour with Implicit-Explicit Learning Framework for Legged Robots}

\author{Shixin Luo$^{\dagger1}$, Songbo Li$^{\dagger1}$, Ruiqi Yu$^{1}$, Zhicheng Wang$^{1}$, Jun Wu$^{1,2}$ and Qiuguo Zhu$^{*1,2}$%
\thanks{Manuscript received: April 23, 2024; Revised July 25, 2024; Accepted August 23, 2024.}
\thanks{This paper was recommended for publication by Editor Aleksandra Faust upon evaluation of the Associate Editor and Reviewers' comments.
This work was supported by the National Key R\&D Program of China (Grant No. 2022YFB4701502), the ”Leading Goose” R\&D Program of Zhejiang(Grant No. 2023C01177), the Key R\&D Project on Agriculture and Social Development in Hangzhou City (Asian Games) (Grant No. 20230701A05), and the Key Research Project of Zhejiang Lab (Grant No. 2021NB0AL03).} 
\thanks{$^{1}$The authors are with Institute of Cyber-Systems and Control, Zhejiang University, 310027, China.
    {\tt\footnotesize \{shixin\_luo, 3190105314, yrq, 3160105273\}@zju.edu.cn, jwu@iipc.zju.edu.cn, qgzhu@zju.edu.cn}}%
\thanks{$^{2}$Qiuguo Zhu and Jun Wu are with State Key Laboratory of Industrial Control Technology, 310027, China.}%
\thanks{$^\dagger$Shixin Luo and Songbo Li contributed equally to this work.}%
\thanks{$^*$Qiuguo Zhu is the corresponding author.}%
\thanks{The supplementary video is available at \url{https://youtu.be/XsjFNcND6js?si=9eLiI8P3fTAXH1mc}.}%
\thanks{Digital Object Identifier (DOI): see top of this page.}
}

\markboth{IEEE Robotics and Automation Letters. Preprint Version. Accepted August, 2024}
{Luo \MakeLowercase{\textit{et al.}}: PIE: Parkour with Implicit-Explicit Learning Framework for Legged Robots}

\maketitle

\begin{abstract}
Parkour presents a highly challenging task for legged robots, requiring them to traverse various terrains with agile and smooth locomotion. This necessitates comprehensive understanding of both the robot's own state and the surrounding terrain, despite the inherent unreliability of robot perception and actuation. Current state-of-the-art methods either rely on complex pre-trained high-level terrain reconstruction modules or limit the maximum potential of robot parkour to avoid failure due to inaccurate perception. In this paper, we propose a one-stage end-to-end learning-based parkour framework: Parkour with Implicit-Explicit learning framework for legged robots (PIE) that leverages dual-level implicit-explicit estimation. With this mechanism, even a low-cost quadruped robot equipped with an unreliable egocentric depth camera can achieve exceptional performance on challenging parkour terrains using a relatively simple training process and reward function. While the training process is conducted entirely in simulation, our real-world validation demonstrates successful zero-shot deployment of our framework, showcasing superior parkour performance on harsh terrains.
\end{abstract}

\begin{IEEEkeywords}
Legged robots, reinforcement learning, deep learning for visual perception.
\end{IEEEkeywords}

%
\IEEEpeerreviewmaketitle

\section{Introduction}

\IEEEPARstart{P}{arkour}, an extremely challenging and exhilarating sport, encompasses a variety of movements such as running, jumping and climbing to traverse obstacles and terrain as quickly and smoothly as possible. Integrating quadruped robots into parkour entails overcoming numerous technical difficulties~\cite{hoeller2024anymal, zhuang2023robot, cheng2023extreme}. For robots, parkour requires a high degree of balance and stability to avoid falls or loss of control while moving rapidly and performing jumps. With obstacles and terrains constantly changing, robots need to accurately perceive their surroundings in real-time and make quick and precise adjustments to avoid collisions or falls. However, successful implementation could open up new possibilities for robot technology in extreme environments and drive advancements in real-world applications and development.

Recent advancements in learning-based approaches have demonstrated significant strides in blind locomotion. Leveraging robust proprioceptive sensors, a quadruped robot can perform state estimation or system identification, achieving promising performance in walking, climbing stairs and mimicking animals~\cite{hwangbo2019learning, nahrendra2023dreamwaq, ji2022concurrent, peng2020learning}. Although the robot can implicitly infer its surrounding terrains, it is still too difficult for a blind policy to foresee the upcoming dangers posed by the harsh terrains and make appropriate response immediately in the extreme tasks like parkour.

To tackle the complexities of varied terrain environments and accomplish more intricate tasks, exteroceptive sensors such as cameras and LiDARs have been employed, yielding advancements in various scenarios including locomotion and manipulation~\cite{miki2022learning, cheng2023legs, agarwal2023legged, yang2021learning, yang2023neural}. Recent studies indicate that even a low-cost quadruped robots equipped with a depth camera can achieve commendable results in parkour~\cite{cheng2023extreme}.

\begin{figure}[t]
\centering
\includegraphics[width=\columnwidth]{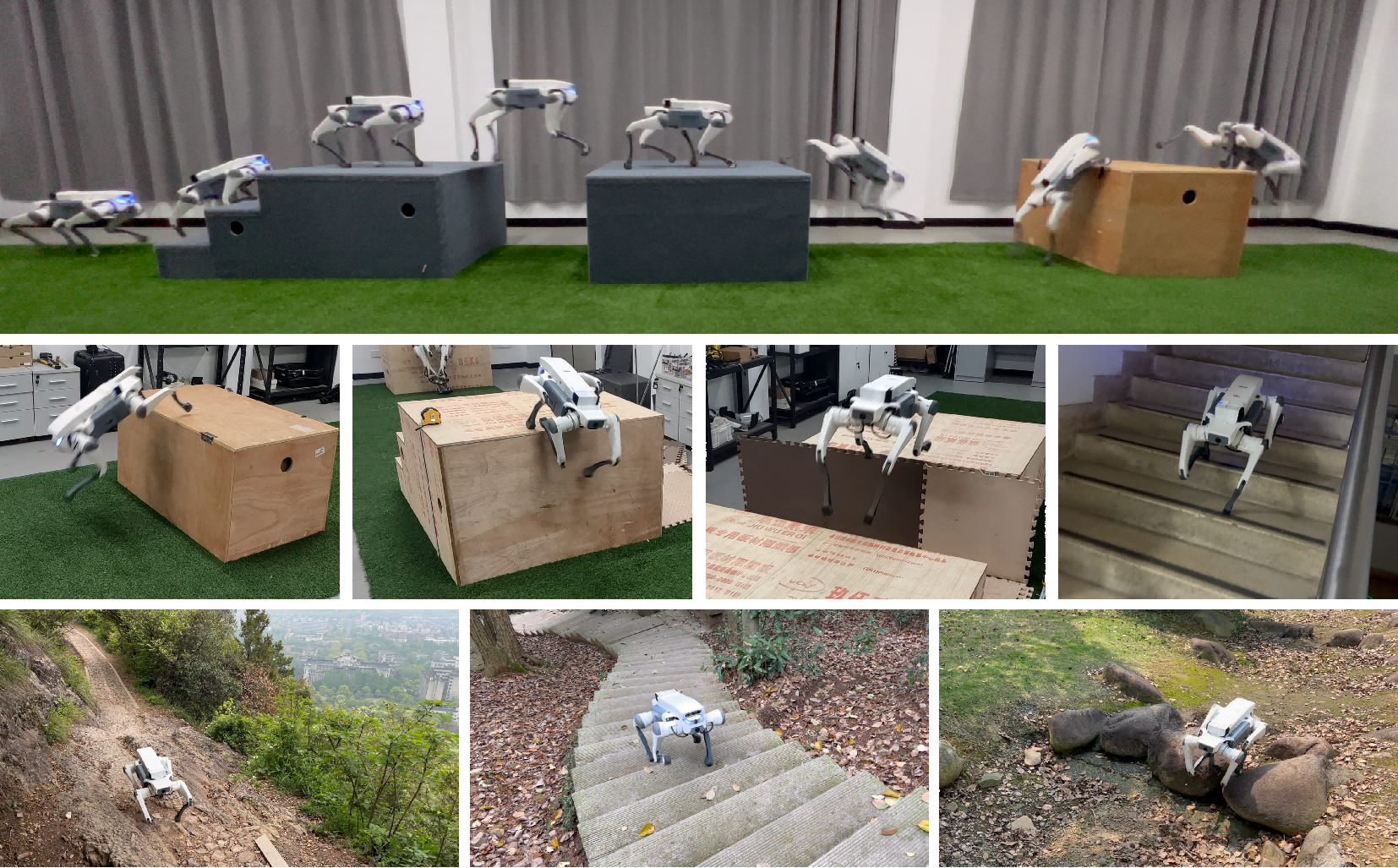}
\caption{Our parkour framework with implicit-explicit estimation allows low-cost robots to traverse a series of challenging parkour-like terrains. Notably, in the real world the robot is not limited to the terrain on which it was trained in simulation, demonstrating impressive generalization capabilities.}
\label{fig_1}
\end{figure}

However, several challenges persist, limiting the maximal performance of quadruped robots in parkour: Firstly, exteroceptive sensors, although essential for enhancing performance, suffer from latency and noise, which poses challenges in achieving real-time and precise terrain estimation, crucial for maximizing robot performance in parkour where robots frequently encounter edges and must avoid falls. Secondly, existing learning-based parkour approaches usually adopt a two-stage training paradigm, complicating the training process and potentially leading to information loss during mimicking learning phases, whether achieved through imitation learning or adaptation methods. Thirdly, integrating multiple behaviors seamlessly into a neural network via a relatively simple training process and reward function remains challenging. To address all these challenges, can we enhance the robot's understanding of its own state and surroundings through a simple one-stage training framework, pushing the limits of parkour further?

In this paper, we propose a one-stage end-to-end reinforcement learning-based framework: Parkour with Implicit-Explicit learning framework for legged robots~(PIE), which elevates the parkour capabilities of quadruped robots to the next level through a simple yet effective training process. Motivated by recent works~\cite{cheng2023extreme, nahrendra2023dreamwaq}, we introduce a hybrid dual-level implicit-explicit estimation approach. Firstly, on the level of understanding the robot's state and surroundings, PIE goes beyond the explicit understanding of the terrain through exteroceptive sensors. Instead, it integrates real-time and robust proprioception with exteroception to implicitly infer the robot's state and surroundings by estimating its successor state, which significantly enhances estimation accuracy. Furthermore, concerning whether the vector representing the robot's state and surroundings is an encoded latent vector or an explicit physical quantity, it explicitly estimates specific physical quantities together with latent vectors, which greatly enhances the quadruped robot's performance in executing parkour maneuvers on challenging terrains. We deployed PIE on a low-cost DEEP Robotics Lite3 quadruped robot and conducted various tests in parkour terrains and real-world scenarios, demonstrating remarkable performance and robustness.

In summary, our main contributions include:
\begin{itemize}
\item{A novel one-stage learning-based parkour framework utilizing dual-level implicit-explicit estimation is proposed to enhance the quality of estimating the robot’s state and surroundings.}
\item{Experiments were conducted to show that our framework enables the robot to leap onto and jump off steps 3x its height, negotiate gaps 3x its length, and climb up and down stairs 1x its height. These results push the limits of parkour for the quadruped robot to a new level.}
\item{Our real-world validation demonstrates effective zero-shot deployment of our framework without extensive fine-tune in the indoor environment and in the wild, showcasing great robustness of our sim-to-real transfer achieved through an elegant and potent pipeline.}
\end{itemize}

\section{Related Work}
\label{chap:2}

\subsection{Vision-Guided Locomotion}

Vision-guided locomotion plays a pivotal role in enhancing the autonomy and adaptability of legged robots. Traditional approaches typically decouple this problem into two components: perception and controller. The perception component translates visual inputs into elevation maps or traversability maps to guide the robot's locomotion. Meanwhile, the controller component employs either model-based methods~\cite{fankhauser2014robot, fankhauser2018probabilistic} or RL methods~\cite{miki2022learning, gangapurwala2021real}. Some approaches bypass explicit map construction, wherein both perception and controller utilize learning-based techniques to generate hierarchical RL methods~\cite{jain2020pixels, kareer2023vinl}. Most of these methods are primarily used for robot navigation and obstacle avoidance, rather than flexible adaptation to complex terrains, as the decoupling process ultimately results in information loss and system delays. Recently, vision-guided RL methods have generally employed end-to-end control systems, showing great promise in traversing complex terrains. For example, Agarwal \textit{et al.} \cite{agarwal2023legged} designed a two-stage learning framework where a student policy, guided by a teacher policy with privileged information inputs, directly predicts joint angles based on depth camera inputs and proprioceptive feedback. Yang \textit{et al.} \cite{yang2021learning} proposed a coupled training framework utilizing a transformer structure to integrate both proprioception and visual observations. Leveraging a self-attention mechanism, it fuses these inputs, enabling autonomous navigation in indoor and outdoor environments with varying obstacles. Yang \textit{et al.} \cite{yang2023neural} utilized a 3D voxel representation with $SE(3)$ equivariance as the extracted feature from visual inputs, achieving precise understanding of terrain.

\begin{figure*}[t]
\centering
\includegraphics[width=6in]{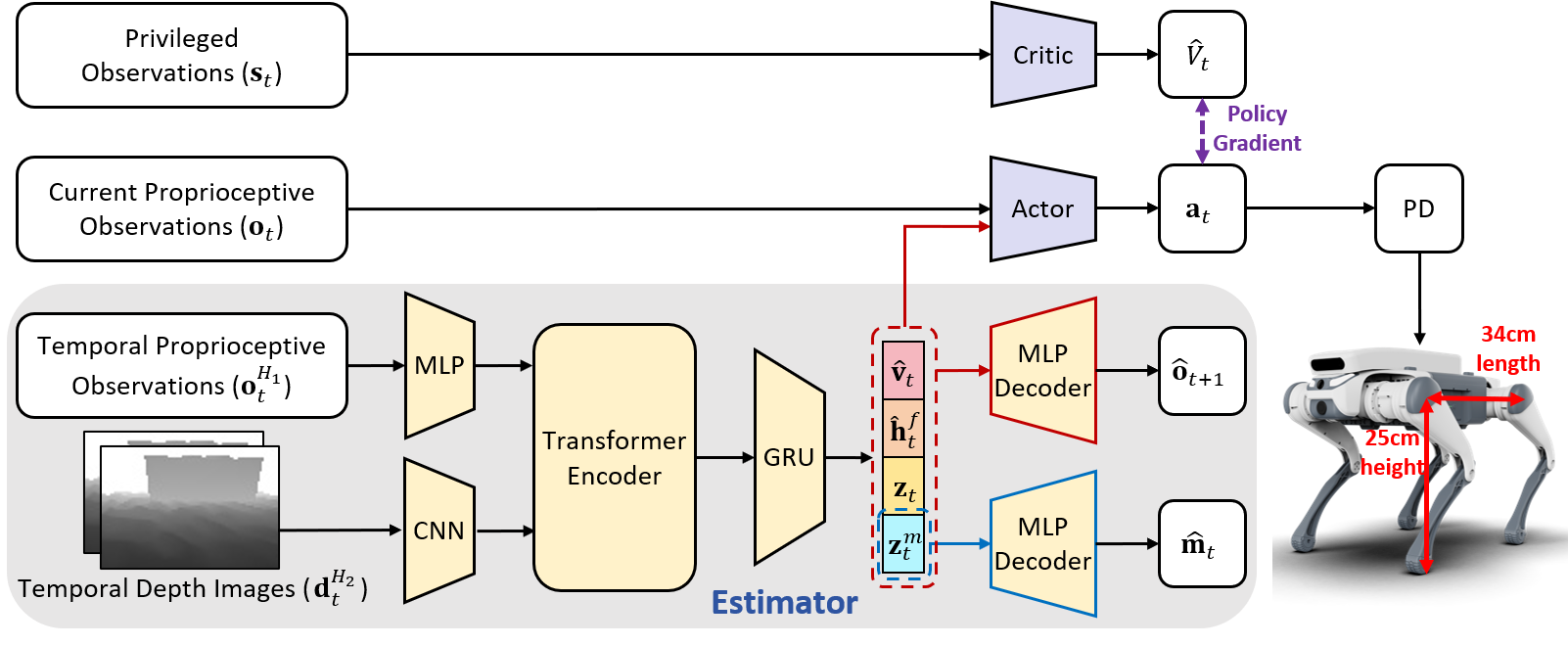}
\caption{Overview of the proposed PIE framework. The grey box represents the estimator, which utilizes implicit-explicit estimation to provide the policy network with the estimated vectors. The framework is concurrently optimized with PPO for the actor and critic network and regression for the estimator.}
\label{fig_2}
\end{figure*}

\subsection{Robot Parkour}

In the context of parkour, constructing precise environmental understanding through exteroceptive and proprioceptive sensors in highly dynamic situations to achieve agile, nimble and robust movement is paramount. Hoeller \textit{et al.} \cite{hoeller2024anymal} described a hierarchical pipeline for navigation in parkour terrains. However, when encountering unstructured terrain, occupancy voxels obtained from an encoder-decoder architecture often imply incorrectness and mistrust, leading to inappropriate responses from navigation and locomotion modules, along with high training costs and low scalability. Zhuang \textit{et al.} \cite{zhuang2023robot} proposed a multi-stage method leveraging soft/hard constraints to accelerate the training process, enabling the robot to learn traversing various terrains directly from depth images. However, its privileged physics information is strongly associated with the geometric properties of obstacles in simulation. This makes it challenging for training the robot to cope with terrain that cannot be described solely by geometric properties. Cheng \textit{et al.} \cite{cheng2023extreme} adopted a framework akin to~\cite{agarwal2023legged}. Differently, it introduces waypoints into the teacher policy's privileged inputs to guide the student policy to autonomously learn heading, but it is the manual specification of waypoints based on terrain that imposes considerable limitations on this method. Futhermore, the two-stage training paradigm used by all aforementioned parkour works results in information loss and performance degradation of the deployed student policy. Meanwhile, these works primarily focus on explicit terrain estimation distilled from the teacher policy, and lack the implicit estimation of the surroundings, thereby hindering the robot's maximal performance capabilities.

\section{Method}
\label{chap:3}

The proposed PIE framework is a one-stage end-to-end learning-based framework that utilizes a single neural network to directly derive desired joint angle commands from raw depth images and onboard proprioception. It circumvents the performance reductions associated with the two-stage training paradigm prevalent in prior learning-based parkour methodologies, aiming to enhance the estimation effectiveness of the robot's state and surroundings through dual-level implicit-explicit estimation. Here, we provide an overview of the framework, followed by detailed expositions.

\subsection{Overview}

As shown in Fig.~\ref{fig_2}, the PIE framework comprises three sub-networks: actor, critic and estimator which will be elaborated upon in Section~\ref{chap:3.2}. An asymmetric actor-critic architecture is adopted to simplify the two-stage training paradigm into a single stage~\cite{nahrendra2023dreamwaq}. The actor network receives inputs solely from proprioception and exteroception data obtainable during deployment, whereas the critic network can incorporate additional privileged information. The optimization process consists of two concurrent parts: the optimization of the actor-critic and the optimization of the estimator. In this work, the actor-critic is optimized using the proximal policy optimization (PPO) algorithm~\cite{schulman2017proximal}.

\subsubsection{Policy Network}

The policy network takes the proprioceptive observation $\mathbf{o}_{t}$, the estimated base velocity $\hat{\mathbf{v}}_{t}$, the estimated foot clearance $\hat{\mathbf{h}}_{t}^{f}$, the encoded height map estimation $\mathbf{z}_{t}^{m}$ and the purely latent vector $\mathbf{z}_{t}$ as input, producing action $\mathbf{a}_{t}$ as output. The proprioceptive observation $\mathbf{o}_{t}$ is a 45-dimensional vector directly measured from joint encoders and the IMU, which is defined as 
\begin{equation}
  \mathbf{o}_t=\begin{bmatrix}\boldsymbol{\omega}_{t} & \mathbf{g}_{t} & \mathbf{c}_{t} & \boldsymbol{\theta}_{t} & \dot{\boldsymbol{\theta}}_{t} & \mathbf{a}_{t-1} \end{bmatrix}^T,
\end{equation}
encompassing the body angular velocity $\boldsymbol{\omega}_{t}$, the gravity direction vector in the body frame $\mathbf{g}_{t}$, the velocity command $\mathbf{c}_{t}$, the joint angle $\boldsymbol{\theta}_{t}$, the joint angular velocity $\dot{\boldsymbol{\theta}}_{t}$ and the previous action $\mathbf{a}_{t-1}$. $\hat{\mathbf{v}}_{t}$, $\hat{\mathbf{h}}_{t}^{f}$, $\mathbf{z}_{t}^{m}$ and $\mathbf{z}_{t}$ are the outputs of the estimator.

\subsubsection{Value Network}

To obtain a more accurate estimation of the state value $\hat{\mathbf{v}}_{t}$, the value network's input $\mathbf{s}_{t}$ includes not only the proprioceptive observation $\mathbf{o}_{t}$ but also the privileged observations base velocity $\mathbf{v}_{t}$ and height map scan dots $\mathbf{m}_{t}$, which is defined as
\begin{equation}
  \mathbf{s}_{t}=\begin{bmatrix}\mathbf{o}_{t} & \mathbf{v}_{t} & \mathbf{m}_{t}\end{bmatrix}^T.
\end{equation}

\subsubsection{Action Space}

The action $\mathbf{a}_{t}$ is a 12-dimensional vector corresponding to the 12 joints of the quadruped robot. For the stability of the policy network output, the action $\mathbf{a}_{t}$ is added as bias to the robot's standstill pose $\boldsymbol{\theta}_{stand}$. Thus, the final target joint angle  $\boldsymbol{\theta}_{target}$ is defined as 
\begin{equation} 
\boldsymbol{\theta}_{target} = \boldsymbol{\theta}_{stand} + \mathbf{a}_{t}.
\end{equation}

\subsubsection{Reward Function}

To underscore the robust performance of our framework, we utilize a relatively simple reward function in a parkour scenario, closely aligned with prior research on blind walking~\cite{rudin2022learning, lee2020learning}. Details of the reward function are provided in Table~\ref{table:1}. $\mathbf{g}$, $\boldsymbol{\tau}$ and $n_{collision}$ are the gravity vector in the body frame, joint torque and the number of collision points other than the feet, respectively.

\begin{table}[t]
    \caption {REWARD FUNCTION ELEMENTS}
    \label{table:1}
    \centering
    \begin{tabular}{ccc}
        \toprule
        Reward & Equation($r_i$) & Weight($w_i$)\\
        \midrule
        Lin. velocity tracking & $\exp\{-4(\mathbf{v}^\mathrm{cmd}_{xy}-\mathbf{v}_{xy})^{2}\}$ & $1.5$ \\ 
        Ang. velocity tracking & $\exp\{-4(\omega^\mathrm{cmd}_\mathrm{yaw}-\omega_\mathrm{yaw})^{2}\}$ & $0.5$ \\ 
        Linear velocity ($z$) & $v^2_z$ & $-1.0$ \\ 
        Angular velocity ($xy$) & $\boldsymbol{\omega}^2_{xy}$ & $-0.05$ \\ 
        Orientation & $|\mathbf{g}|^2$ & $-1.0$ \\ 
        Joint accelerations &  $\boldsymbol{\ddot{\theta}}^2$ &  $-2.5\times10^{-7}$ \\ 
        Joint power & $|\boldsymbol{\tau}||\boldsymbol{\dot{\theta}}|$ & $-2\times10^{-5}$ \\
        Collision & $-n_{collision}$ & $-10.0$ \\ 
        Action rate & $(\mathbf{a}_t - \mathbf{a}_{t-1})^2$ & $-0.01$ \\ 
        Smoothness & $(\mathbf{a}_t - 2\mathbf{a}_{t-1} + \mathbf{a}_{t-2})^2$ & $-0.01$ \\ 
        \bottomrule
    \end{tabular} 
\end{table}

\subsection{Estimator}
\label{chap:3.2}

The vectors to be estimate, $\hat{\mathbf{v}}_{t}$, $\hat{\mathbf{h}}_{t}^{f}$, $\mathbf{z}_{t}^{m}$ and $\mathbf{z}_{t}$, can be divided into implicit and explicit estimations on two levels.

On the level of understanding the robot's state and surroundings, a blind robot can implicitly infer its state and surrounding terrain solely through robust onboard sensors by estimating the successor proprioceptive state~\cite{nahrendra2023dreamwaq, long2023hybrid}. However, it must interact with the environment and then adjust its inference based on immediate feedback. In the parkour scenario, the robot must anticipate the terrain ahead. For example, in high or long jumps, the robot needs to gather momentum in advance to execute liftoff successfully, which can only be achievable through exteroceptive sensors providing additional information. However, solely explicitly estimating terrain using laggy and noisy depth cameras is not sufficiently reliable, especially near edges where the risk of stepping off is high~\cite{zhuang2023robot, cheng2023extreme}. Hence, we propose a multi-head auto-encoder mechanism to integrate implicit and explicit estimation of the robot's state and surroundings.

The encoder module has two inputs: temporal depth images $\mathbf{d}_{t}^{H_{2}}$ and temporal proprioceptive observations $\mathbf{o}_{t}^{H_{1}}$. The depth observations in the near past are stacked in the channel dimension to form $\mathbf{d}_{t}^{H_{2}}$, while the proprioceptive observations in the near past are concatenated to form $\mathbf{o}_{t}^{H_{1}}$. In this work, we set $H_{1} = 10$ and $H_{2} = 2$. $\mathbf{d}_{t}^{H_{2}}$ is processed by a CNN encoder to extract depth feature maps, while $\mathbf{o}_{t}^{H_{1}}$ is processed by an MLP encoder to extract proprioceptive features. To facilitate cross-modal reasoning between visual tokens from the 2D depth feature map and proprioceptive features, we employ a shared transformer encoder to further integrate the two modalities. Since the robot is equipped with only one egocentric depth camera, it cannot directly perceive terrain beneath or behind its body visually. Therefore, the outputs of the transformer, which are the depth and proprioceptive features after cross-modal attention processing, are concatenated and fed into a GRU to generate memories of the state and terrain. The outputs of the GRU are then served as the policy network's input vectors. One of these vectors is a encoded height map estimation vector $\mathbf{z}_{t}^{m}$, which is decoded into $\hat{\mathbf{m}}_{t}$ to reconstruct the high-dimensional height map $\mathbf{m}_{t}$, serving as the explicit estimation of the surrounding terrain. Another vector is a purely latent vector $\mathbf{z}_{t}$. Instead of using a pure auto-encoder, following prior work~\cite{nahrendra2023dreamwaq}, we adopt a VAE structure to extract features for $\mathbf{z}_{t}$. Namely, we employ the KL divergence as the latent loss. When decoded alongside other vectors with explicit meanings, it reconstructs $\hat{\mathbf{o}}_{t+1}$ to represent the robot’s successor state $\mathbf{o}_{t+1}$, encapsulating implicit information about the robot's state and surroundings. We use mean-squared-error (MSE) loss between the decoded reconstruction and the ground truth for both explicit and implicit estimations. This constitutes the first level of implicit-explicit estimation.

\begin{figure*}[!t]
\centering
\includegraphics[width=\textwidth]{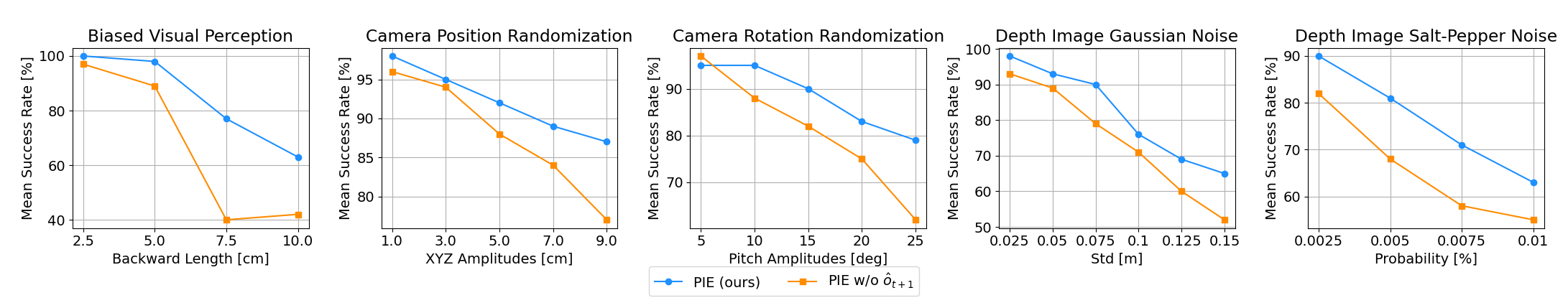}
\caption{Simulation experiments results for PIE and PIE without $\hat{\mathbf{o}}_{t+1}$ in the presence of various camera input errors. Five plots represent the five types of camera errors introduced, with the x-axis of each plot representing: the distance deviates of the depth image from the actual terrain perception in the world coordinate system along the negative x-axis; the maximum value of uniformly distributed noise added to the position of the camera relative to the base link of the robot in the x, y and z directions; the maximum value of uniformly distributed noise added to the pitch of the camera coordinate system relative to the robot coordinate system; the standard deviation of Gaussian noise added to the depth image; the amount of salt-pepper noise added to the depth image, with random pixels being set to either the minimum or maximum depth based on the probability. These experiments adopt the highest difficulty level of terrains encountered by the robot during training, with the y-axis representing the average success rate across all terrains, including gap, stairs and step. Similarly, to reflect the generality of the experimental results, the average results of one hundred robots are taken for each experiment.}
\label{fig_3}
\end{figure*}

The second level concerns whether the vector is an encoded latent vector or an explicit physical quantity. $\mathbf{z}_{t}^{m}$ and $\mathbf{z}_{t}$ are encoded latent vectors followed by a decoder for reconstruction. Through this compression and dimension reduction, noise influence is reduced, resulting in greater robustness. For the explicit physical quantities, $\hat{\mathbf{v}}_{t}$ and $\hat{\mathbf{h}}_{t}^{f}$ are explicitly estimated to prioritize velocity tracking during training and provide relevant foot clearance information crucial for understanding terrain in parkour scenarios. The loss for this explicit physical quantities estimation part is also mean-squared-error (MSE) loss. This constitutes the second level of implicit-explicit estimation.

\begin{table}[t]
    \caption {DOMAIN RANDOMIZATION RANGES}
    \label{table:2}
    \centering
    \begin{tabular}{ccc}
        \toprule
        Parameter & Randomization range & Unit\\
        \midrule
        Payload & $[-1, 2]$ &$\mathrm{~kg}$ \\ 
        $K_p$ factor & $[0.9, 1.1]$ & $\mathrm{~Nm/rad}$ \\ 
        $K_d$ factor & $[0.9, 1.1]$ & $\mathrm{~Nms/rad}$ \\ 
        Motor strength factor & $[0.9, 1.1]$ & $\mathrm{~Nm}$ \\ 
        Center of mass shift & $[-50,50]$ & $\mathrm{~mm}$ \\
        Friction coefficient & $[0.2,1.2]$ & - \\
        Initial joint positions & $[0.5,1.5]$ & rad \\
        System delay & $[0,15]$ & $\mathrm{ms}$ \\
        Camera position ($x$) & $[-10,10]$ & $\mathrm{~mm}$ \\
        Camera position ($y$) & $[-10,10]$ & $\mathrm{~mm}$ \\
        Camera position ($z$) & $[-10,10]$ & $\mathrm{~mm}$ \\
        Camera pitch & $[-1,1]$ & $\mathrm{~deg}$ \\
        Camera horizontal FOV & $[86,88]$ & $\mathrm{~deg}$ \\
        \bottomrule
    \end{tabular} 
\end{table}

The overall training loss for our estimator is defined as 
\begin{equation} 
\begin{aligned}
\mathcal{L} = &D_\text{KL}(q(\mathbf{z}_{t}|\mathbf{o}^{{H}_{1}}_{t}, \mathbf{d}^{{H}_{2}}_{t})\parallel p(\mathbf{z}_{t})) + \text{MSE}(\hat{\mathbf{o}}_{t+1}, \mathbf{o}_{t+1})\\
&+ \text{MSE}(\hat{\mathbf{m}}_{t}, \mathbf{m}_{t}) + \text{MSE}(\hat{\mathbf{v}}_{t}, \mathbf{v}_{t}) + \text{MSE}(\hat{\mathbf{h}}_{t}^{f}, \mathbf{h}_{t}^{f}),
\end{aligned}
\end{equation}
where $q(\mathbf{z}_{t}|\mathbf{o}^{{H}_{1}}_{t}, \mathbf{d}^{{H}_{2}}_{t})$ is the posterior distribution of the $\mathbf{z}_{t}$, given $\mathbf{o}^{{H}_{1}}_{t}$ and $\mathbf{d}^{{H}_{2}}_{t}$. $p(\mathbf{z}_{t})$ is the prior distribution of $\mathbf{z}_{t}$ parameterized by a standard normal distribution.

\subsection{Training Details}

\subsubsection{Simulation Platform}

We utilize Isaac Gym with 4096 parallel environments to train the actor, critic and estimator in simulation. Leveraging NVIDIA Warp, we are able to train 10,000 iterations in under 20 hours on the NVIDIA RTX 4090, resulting in a readily deployable network.

\subsubsection{Training Curriculum}

Following the principles of prior works, we adopt a curriculum where the difficulty of the terrain progressively increased to enable the policy to adapt to increasingly challenging environments~\cite{rudin2022learning}. Our parkour terrains include gaps up to 1m wide, steps up to 0.75m height, hurdles up to 0.75m height and stairs with a height up to 0.25m. Lateral velocity commands are sampled from the range of $[0.0, 1.5]$ m/s, and horizontal angular velocity are sampled from the range of $[-1.2, 1.2]$ rad/s.

Additionally, to ensure that the deployed policy can traverse terrains beyond the fixed paradigms in simulation, besides manifesting variations in terrain difficulty, we introduce some randomization to various terrains in simulation. For instance, to enable the robot to adapt to the large gap which can be easily mistaken as a flat ground between two steps where it can jump down and then jump up to the next step, we randomize the depth and width of the gap.

\subsubsection{Domain Randomization}

To enhance the robustness of the network trained in simulation and facilitate smooth sim-to-real transfer, we randomize the mass of the robot body, the center of mass (CoM) of the robot, the mass payload, the initial joint positions, the ground friction coefficient, the motor strength, the PD gains, the system delay, as well as the camera position, orientation and field of view. The ranges for randomization of each parameter are specified in Table~\ref{table:2}.

\section{Experiments}
\label{chap:4}

To evaluate the effectiveness of our framework, we conducted ablation studies in both simulation and real-world environments, and compared them with existing parkour frameworks. Here, we describe our experimental setup, present the experimental results and provide analysis.

\subsection{Experimental Setup}

We deployed PIE on a DEEP Robotics Lite3, which is a low-cost quadruped robot. As illustrated in Fig.~\ref{fig_2}, when standing, the thigh joint height is 25cm, and the distance between the two thigh joints measures 34cm. The Lite3 weighs 12.7kg, with a peak knee joint torque of 30.5Nm~\cite{deeprobotics2024lite3, deeprobotics2024joint}. Similarly, the Unitree A1, used by Zhuang~\textit{et al.}~\cite{zhuang2023robot} and Cheng~\textit{et al.}~\cite{cheng2023extreme}, has a thigh joint height of 26cm and a distance of 40cm between the thigh joints~\cite{unitree2024a1}. It weighs 12kg, with a peak knee joint torque of 33.5Nm. Thus, it is reasonable to use their results as baselines in our comparative experiments. Raw depth images are received at a frequency of 10Hz using its onboard Intel RealSense D435i camera. Joint actions are outputted at a frequency of 50Hz. Image post-processing and network inference are performed using its onboard Rockchip RK3588 processor.

We designed five sets of ablations to compare against ours, aiming to verify the effectiveness of PIE's dual-level implicit-explicit estimation:

\begin{itemize}
\item{\textit{PIE w/o reconstructing $\hat{\mathbf{o}}_{t+1}$}: This method can only estimate the robot's state and surroundings explicitly without reconstructing $\mathbf{o}_{t+1}$.}
\item{\textit{PIE w/o reconstructing $\hat{\mathbf{m}}_{t}$}: This method utilizes both proprioception and exteroception but only reconstructs $\mathbf{o}_{t+1}$ as implicit estimation of surrounding, lacking explicit estimation of terrain.}
\item{\textit{PIE w/o estimating $\hat{\mathbf{v}}_{t}$}: This method is trained without estimating $\hat{\mathbf{v}}_{t}$.}
\item{\textit{PIE w/o estimating $\hat{\mathbf{h}}_{t}^{f}$}: This method is trained without estimating $\hat{\mathbf{h}}_{t}^{f}$.}
\item{\textit{PIE using predicted $\mathbf{o}_{t+1}$}: This method allows the policy network to use the predicted $\mathbf{o}_{t+1}$ as input, rather than the pure latent vector $\mathbf{z}_{t}$ used for reconstruction.}
\end{itemize}

\subsection{Simulation Experiments}

In simulation, we implemented an environment similar to the training setup for metric comparison, wherein we evaluated the performance of PIE against other ablations on different terrains. To precisely assess the policy's traversal capability across terrains, including gap, stairs and step (hurdle was not specifically tested in simulation as the randomized step lengths covered the hurdle characteristics), we configured curriculum-like terrains with ten levels of increasing difficulty per column. We measured average level of terrain difficulty that robots can reach by letting robots start from the easiest level and gradually traversing the rest of their column until termination due to falls or collisions. For each terrain type, we created forty sets of environments, with one hundred robots tested simultaneously to ensure randomness in terrains and robots.

\begin{table}[t]
    \caption {ABLATION STUDY FOR THE VECTORS TO BE ESTIMATED:\\
                PERFORMANCE IN SIMULATION}
    \label{table:3}
    \centering
    \begin{tabular}{>{\centering\arraybackslash}m{2cm}*{3}{>{\centering\arraybackslash}m{1cm}}}
        \toprule
        \multirow{2}{*}{Method} &
        \multicolumn{3}{c}{Mean Terminated Difficulty Level} \\  
        \cmidrule(lr){2-4}
        & Gap & Stairs & Step \\ 
        \midrule
        \textbf{PIE (ours)} & \textbf{9.9} & \textbf{9.86} & \textbf{9.81} \\ 
        PIE w/o $\hat{\mathbf{o}}_{t+1}$ & 9.51 & 9.45 & 9.62 \\
        PIE w/o $\hat{\mathbf{h}}_{t}^{f}$ & 7.41 & 7.36 & 3.09 \\
        PIE w/o $\hat{\mathbf{v}}_{t}$ & 8.7 & 8.22 & 8.48 \\
        PIE w/o $\hat{\mathbf{m}}_{t}$ & 9.75 & 4.25 & 1.67 \\
        PIE using predicted $\hat{\mathbf{o}}_{t+1}$ & 9.23 & 7.28 & 3.29 \\
        \bottomrule
    \end{tabular} 
\end{table}

\begin{figure*}[t]
\centering
\hspace*{-1cm}
\includegraphics[width=0.95\textwidth]{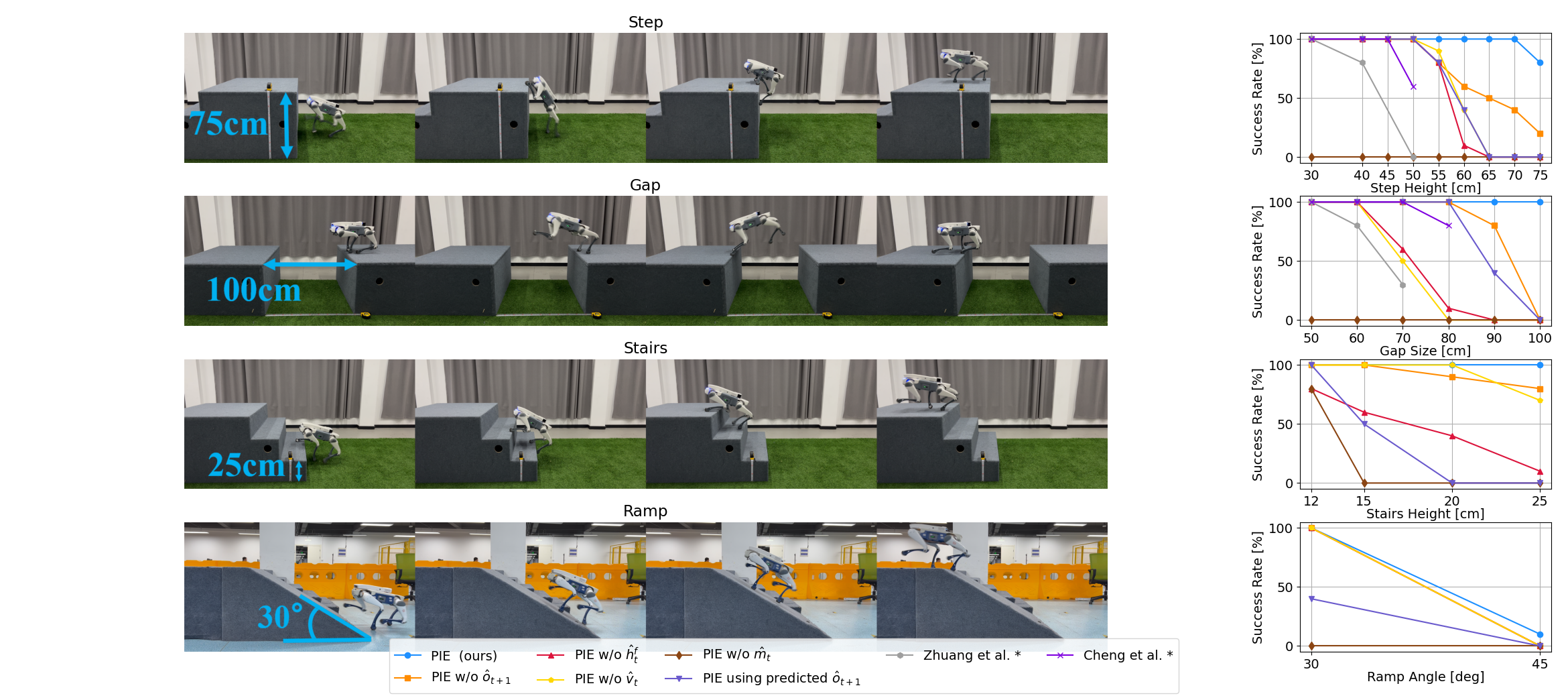}
\caption{Real-world indoor experiments results of each method. We measured the success rates for ten trails on each terrain of each difficulty level. Our PIE framework maintains consistent performance between real-world deployment and simulation, outperforming all other methods. Additionally, it demonstrates a certain level of generalization ability to traverse ramp terrains. Starred are previous related works~\cite{zhuang2023robot, cheng2023extreme}.}
\label{fig_4}
\end{figure*}

As shown in Table~\ref{table:3}, our PIE framework utilizing both implicit and explicit estimation demonstrates superior performance across all terrains. Although PIE w/o $\hat{\mathbf{o}}_{t+1}$ exhibits commendable performance in simulation, its reliance solely on explicit terrain estimation leads to a lack of comprehensive terrain understanding, making it susceptible to falls due to minor deviations in foothold positioning when encountering the most challenging terrains. The performance of PIE w/o $\hat{\mathbf{h}}_{t}^{f}$  significantly declines compared to the former two methods because directly estimating $\hat{\mathbf{h}}_{t}^{f}$ remains crucial for the robot's direct comprehension of the terrain beneath its feet, lacking which can hinder the robot's ability to execute extreme parkour maneuvers by stepping along the terrain edges. Without estimating the robot's velocity, the PIE w/o $\hat{\mathbf{v}}_{t}$ evidently introduces significant biases in velocity tracking, consequently diminishing the robot's maneuverability on high-difficulty terrains. The absence of regression between predicted $\hat{\mathbf{m}}_{t}$ and the ground truth in PIE w/o $\hat{\mathbf{m}}_{t}$ makes it difficult to extract useful terrain information directly from the input depth image, resulting in poor performance on highly challenging parkour terrains. PIE using predicted $\mathbf{o}_{t+1}$ struggles on step and stairs terrain because unlike $\hat{\mathbf{z}}_{t}$, which has a prior distribution parameterized by a standard normal distribution, the distribution of $\mathbf{o}_{t+1}$ is more complex, making it more challenging for the policy network to extract useful information from it.

Additionally, we evaluated the performance of both PIE and PIE w/o $\hat{\mathbf{o}}_{t+1}$ across all terrains in the presence of various camera input errors by employing significantly higher noise levels to the camera, beyond those used in standard randomization. These errors normally cause discrepancies between visual inputs and the actual environment, which often occurs under conditions where the unreliable depth camera is subject to various interference. We employed these metrics to assess the effectiveness of learning implicit-explicit estimation in such scenarios as shown in Fig.~\ref{fig_3}. Overall, PIE outperforms PIE w/o $\hat{\mathbf{o}}_{t+1}$ in both metrics, particularly when considerable noise is added to the camera, resulting in significant performance disparities between the two. In contrast, within the normal domain randomization range, their performances are comparable. This suggests that not only can our learning framework better acquire robot state estimation and terrain understanding, but also, through the estimation of $\hat{\mathbf{o}}_{t+1}$, the policy can make correct decisions when vision contradicts proprioception, placing more trust in proprioception to achieve better performance.

\subsection{Real-World Indoor Experiments}

We compared the success rates of PIE with those of the ablations in the real world. Each method was tested for ten trails on each terrain of each difficulty level, including step, gap and stairs. Additionally, although we did not train specifically on ramp terrains, we still conducted tests on ramp terrains to evaluate the generalization performance. 

\begin{figure*}[t]
\centering
\includegraphics[width=\textwidth]{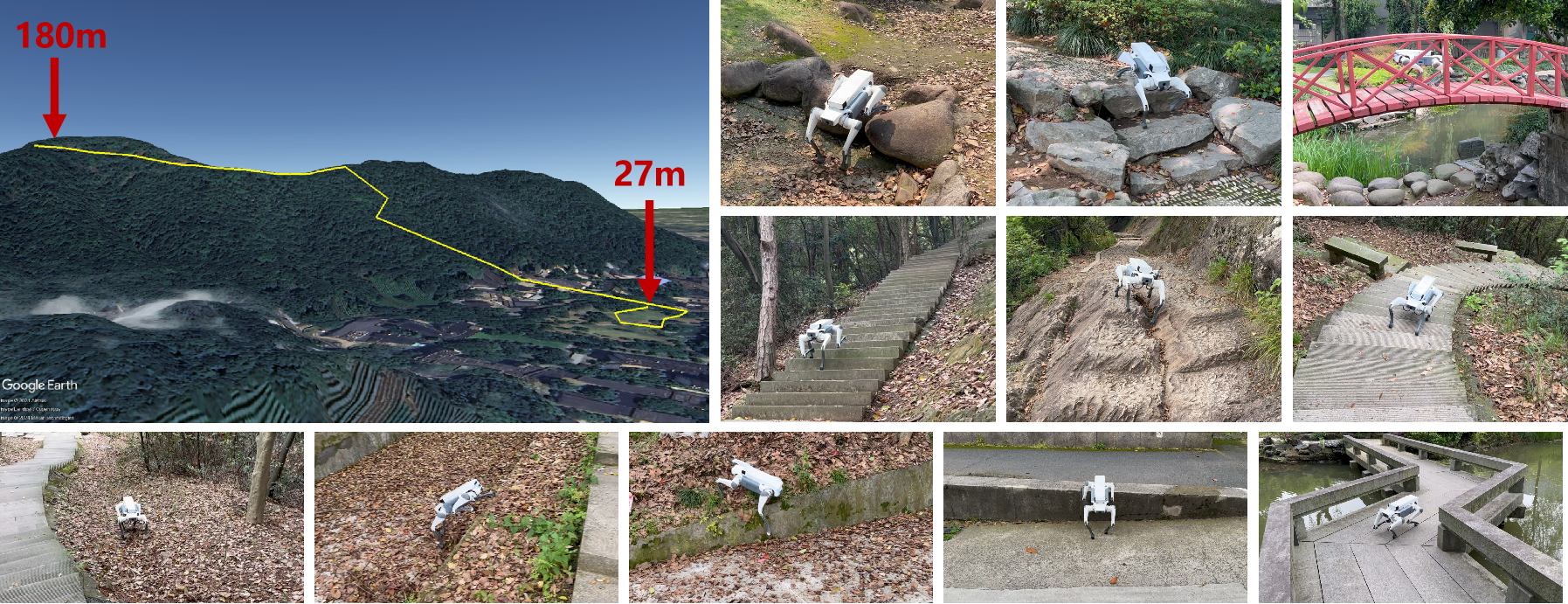}
\caption{A 2km round-trip hike from the ZJU Yuquan campus to the top of Laohe Mountain and back, with
an elevation gain of 153 meters from 27m to 180m. Our PIE framework successfully navigated the challenging terrains along this route, demonstrating notable generalization capabilities.}
\label{fig_5}
\end{figure*}

\begin{figure}[t]
\centering
\includegraphics[width=\columnwidth]{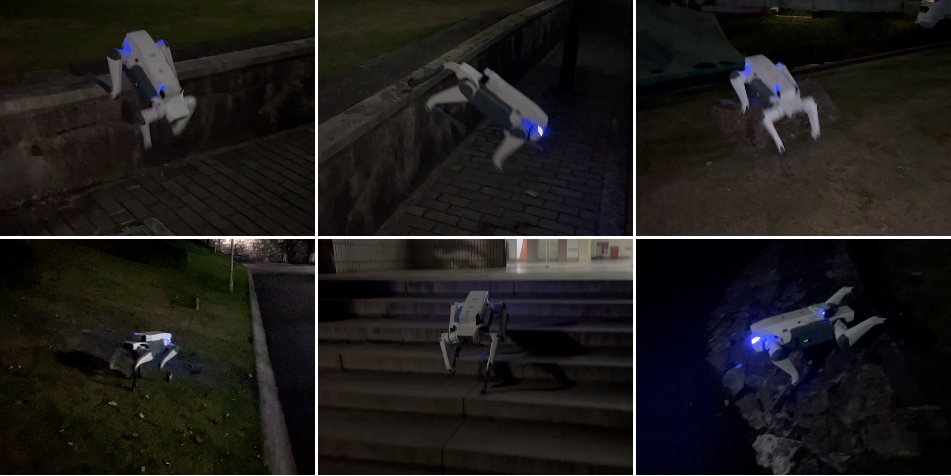}
\caption{Tests in dark outdoor environment. Despite the near absence of visible light, the robot was able to accurately perform agile maneuvers.}
\label{fig_6}
\end{figure}

\begin{table}[t]
    \caption {COMPARISON OF PARKOUR ABILITIES}
    \label{table:4}
    \centering
    \begin{tabular}{ccccc}
        \toprule
        Method & Robot & Step & Gap & Stairs \\
        \midrule
        Hoeller~\textit{et al.}~\cite{hoeller2024anymal} & AnymalC & 2$\times$ & 1.5$\times$ & 0.5$\times$ \\ 
        Zhuang~\textit{et al.}~\cite{zhuang2023robot} & Unitree-A1 & 1.6$\times$ & 1.5$\times$ & - \\
        Cheng~\textit{et al.}~\cite{cheng2023extreme} & Unitree-A1 & 2$\times$ & 2$\times$ & - \\
        \textbf{PIE~(ours)} & DEEP Robotics Lite3 & \textbf{3$\times$} & \textbf{3$\times$} & \textbf{1$\times$} \\
        \bottomrule
    \end{tabular} 
\end{table}

As shown in Fig.~\ref{fig_4} and the supplementary video, owing to the implicit-explicit estimation, our framework shows outstanding performance across all skills, surpassing all other ablations and previous related works. We find that our PIE framework enables the robot to climb obstacles as high as 0.75m (3x robot height), leap over gaps as large as 1m (3x robot length) and climb stairs as high as 0.25m (1x robot height). This represents a significant performance improvement of at least 50\% compared to the state-of-the-art robot parkour frameworks as shown in Table~\ref{table:4}, which indicates the relative size of traversable obstacles with respect to quadruped’s height and length respectively. Notably, PIE exhibits consistent success rates compared to in simulation, demonstrating remarkable sim-to-real transferability. It is worth mentioning that, despite the lack of specific training for ramp terrains in simulation, our framework still demonstrates better generalization performance.

Among the ablations, PIE w/o $\hat{\mathbf{o}}_{t+1}$ performs relatively better, but due to the larger delay and noise in perception and actuation in the real world, its performance noticeably decreases compared to in simulation. PIE w/o $\hat{\mathbf{h}}_{t}^{f}$ struggles to learn to handle terrain edges correctly due to the lack of intuitive estimation of foot clearance, resulting in lower success rates. As for PIE w/o $\hat{\mathbf{v}}_{t}$, when the terrain difficulty increases, the estimation of base velocity deteriorates, leading to a noticeable decline in success rates when following velocity commands. PIE w/o $\hat{\mathbf{m}}_{t}$ faces challenges extracting useful terrain information from more complicated real-world depth image compared to in simulation, making external perception interference rather than assistance, thus resulting in near-zero success rates across all terrains.

\subsection{Real-World Outdoor Experiments}

Due to the significant disturbances the depth camera faces in outdoor settings, the sim-to-real gap becomes more pronounced. To fully assess the robustness and generalization abilities of the PIE framework in such environments, as shown in Fig.~\ref{fig_5}, we conducted a long-distance hike from the ZJU Yuquan campus to the top of Laohe Mountain and back. The round trip covered 2km, with an elevation gain of 153m. Along this trail, the robot was required to traverse long-range continuous curved stairs of varying heights and widths, irregularly shaped steps and hurdles, ramps with steep inclinations, as well as deformable, slippery ground and rocky surfaces. The robot completed the entire hiking trail in just 40 minutes with no stops, except for moments when the operator and photographer struggled to keep up due to the robot’s rapid ascent on inconsistently high and wide stairs. Additionally, as shown in Fig.~\ref{fig_6}, we conducted tests in dim outdoor conditions at night, where the robot continuously jumped over high steps and irregular rocks, and climbed up and down slopes and stairs. Please refer to the supplementary video for the demonstration of the hike and night tests.

\begin{figure}[t]
\centering
\includegraphics[width=\columnwidth]{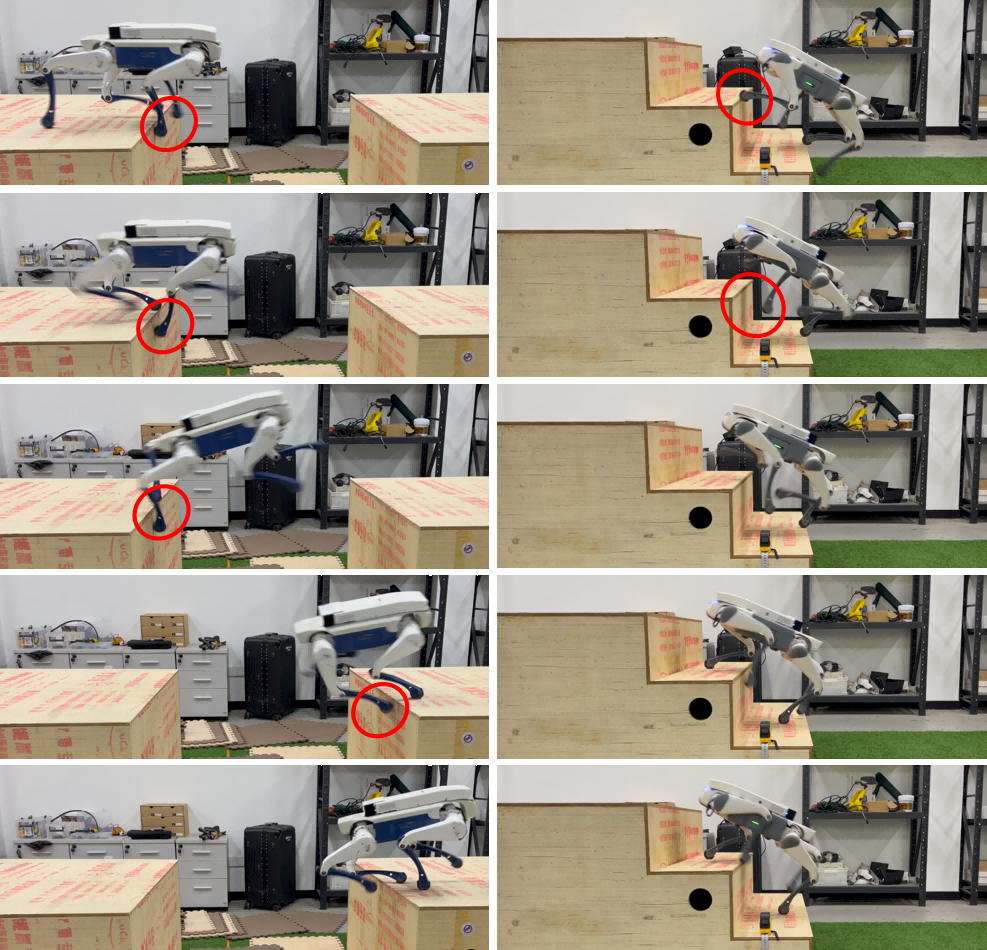}
\caption{The robot can quickly regain stability even when estimations are inaccurate during intense maneuvers. We observed that when leaping over a gap, despite slight distance estimation errors causing the robot's front and rear legs to not fully support on the platform before takeoff, it still managed to execute the jumping action and land smoothly on the other side. Similarly, when encountering a sudden step void while ascending stairs, the robot promptly stabilized itself to continue stepping upwards.}
\label{fig_7}
\end{figure}

\section{Discussion}
\label{chap:5}

Despite not employing imitation learning or designing intricate reward functions to constrain robot behaviors, our straightforward training process enables the robot to achieve a natural and agile gait, seamlessly transitioning across various complex terrains, similar to the natural movements observed in real cats and dogs. We hypothesize that this is attributed to its ability to predict the successor state, which not only enhances its capability to manage delays~\cite{wang2023addressing}, but also establishes an internal model that fosters a better understanding of both itself and its environment~\cite{long2023hybrid}.

Furthermore, in emergent scenarios where slight deviations in external perception occur—such as being momentarily tripped up while running up 25cm stairs at high speed, or misstepping during the takeoff phase of a 1m gap jump, which would typically be catastrophic even for a human traceur—we were surprised to observe that our PIE framework was capable of executing timely and accurate responses, which is shown in Fig.~\ref{fig_7} and the supplementary video. This is particularly noteworthy given that reinforcement learning is generally not adept at handling precise maneuvers, making such deviations difficult to avoid entirely.

However, certain limitations still exist. Firstly, it lacks a 3D understanding of the terrain, thus unable to crouch under obstacles. Secondly, our external perception relies solely on depth images, lacking the richer semantic information provided by RGB images. Additionally, our training is confined to static environments, without extension to dynamic scenes, potentially leading to confusion in visual estimation.

\section{Conclusion}
\label{chap:6}

In this work, we propose a novel one-stage end-to-end learning-based parkour framework PIE, utilizing dual-level explicit-implicit estimation to refine the understanding of both the robot's state and its environment. Compared to existing learning-based parkour frameworks, PIE significantly enhances the parkour capabilities of robots with a unified policy. Meanwhile, PIE demonstrates commendable generalization in outdoor environments, maintaining effective performance across diverse conditions.

In the future, we aim to design a new unified learning-based sensorimotor integration framework, which extracts 3D terrain information from depth images and obtains abundant semantic information from RGB images, to achieve better adaptability and mobility in various environments.

\bibliographystyle{IEEEtran}

\bibliography{references}

\begin{thebibliography}{10}
\providecommand{\url}[1]{#1}
\csname url@samestyle\endcsname
\providecommand{\newblock}{\relax}
\providecommand{\bibinfo}[2]{#2}
\providecommand{\BIBentrySTDinterwordspacing}{\spaceskip=0pt\relax}
\providecommand{\BIBentryALTinterwordstretchfactor}{4}
\providecommand{\BIBentryALTinterwordspacing}{\spaceskip=\fontdimen2\font plus
\BIBentryALTinterwordstretchfactor\fontdimen3\font minus \fontdimen4\font\relax}
\providecommand{\BIBforeignlanguage}[2]{{%
\expandafter\ifx\csname l@#1\endcsname\relax
\typeout{** WARNING: IEEEtran.bst: No hyphenation pattern has been}%
\typeout{** loaded for the language `#1'. Using the pattern for}%
\typeout{** the default language instead.}%
\else
\language=\csname l@#1\endcsname
\fi
#2}}
\providecommand{\BIBdecl}{\relax}
\BIBdecl

\bibitem{hoeller2024anymal}
D.~Hoeller, N.~Rudin, D.~Sako, and M.~Hutter, ``Anymal parkour: Learning agile navigation for quadrupedal robots,'' \emph{Science Robotics}, vol.~9, no.~88, p. eadi7566, 2024.

\bibitem{zhuang2023robot}
Z.~Zhuang, Z.~Fu, J.~Wang, C.~Atkeson, S.~Schwertfeger, C.~Finn, and H.~Zhao, ``Robot parkour learning,'' \emph{arXiv preprint arXiv:2309.05665}, 2023.

\bibitem{cheng2023extreme}
X.~Cheng, K.~Shi, A.~Agarwal, and D.~Pathak, ``Extreme parkour with legged robots,'' \emph{arXiv preprint arXiv:2309.14341}, 2023.

\bibitem{hwangbo2019learning}
J.~Hwangbo, J.~Lee, A.~Dosovitskiy, D.~Bellicoso, V.~Tsounis, V.~Koltun, and M.~Hutter, ``Learning agile and dynamic motor skills for legged robots,'' \emph{Science Robotics}, vol.~4, no.~26, p. eaau5872, 2019.

\bibitem{nahrendra2023dreamwaq}
I.~M.~A. Nahrendra, B.~Yu, and H.~Myung, ``Dreamwaq: Learning robust quadrupedal locomotion with implicit terrain imagination via deep reinforcement learning,'' in \emph{2023 IEEE International Conference on Robotics and Automation (ICRA)}.\hskip 1em plus 0.5em minus 0.4em\relax IEEE, 2023, pp. 5078--5084.

\bibitem{ji2022concurrent}
G.~Ji, J.~Mun, H.~Kim, and J.~Hwangbo, ``Concurrent training of a control policy and a state estimator for dynamic and robust legged locomotion,'' \emph{IEEE Robotics and Automation Letters}, vol.~7, no.~2, pp. 4630--4637, 2022.

\bibitem{peng2020learning}
X.~B. Peng, E.~Coumans, T.~Zhang, T.-W. Lee, J.~Tan, and S.~Levine, ``Learning agile robotic locomotion skills by imitating animals,'' \emph{arXiv preprint arXiv:2004.00784}, 2020.

\bibitem{miki2022learning}
T.~Miki, J.~Lee, J.~Hwangbo, L.~Wellhausen, V.~Koltun, and M.~Hutter, ``Learning robust perceptive locomotion for quadrupedal robots in the wild,'' \emph{Science Robotics}, vol.~7, no.~62, p. eabk2822, 2022.

\bibitem{cheng2023legs}
X.~Cheng, A.~Kumar, and D.~Pathak, ``Legs as manipulator: Pushing quadrupedal agility beyond locomotion,'' in \emph{2023 IEEE International Conference on Robotics and Automation (ICRA)}.\hskip 1em plus 0.5em minus 0.4em\relax IEEE, 2023, pp. 5106--5112.

\bibitem{agarwal2023legged}
A.~Agarwal, A.~Kumar, J.~Malik, and D.~Pathak, ``Legged locomotion in challenging terrains using egocentric vision,'' in \emph{Conference on robot learning}.\hskip 1em plus 0.5em minus 0.4em\relax PMLR, 2023, pp. 403--415.

\bibitem{yang2021learning}
R.~Yang, M.~Zhang, N.~Hansen, H.~Xu, and X.~Wang, ``Learning vision-guided quadrupedal locomotion end-to-end with cross-modal transformers,'' \emph{arXiv preprint arXiv:2107.03996}, 2021.

\bibitem{yang2023neural}
R.~Yang, G.~Yang, and X.~Wang, ``Neural volumetric memory for visual locomotion control,'' in \emph{Proceedings of the IEEE/CVF Conference on Computer Vision and Pattern Recognition}, 2023, pp. 1430--1440.

\bibitem{fankhauser2014robot}
P.~Fankhauser, M.~Bloesch, C.~Gehring, M.~Hutter, and R.~Siegwart, ``Robot-centric elevation mapping with uncertainty estimates,'' in \emph{Mobile Service Robotics}.\hskip 1em plus 0.5em minus 0.4em\relax World Scientific, 2014, pp. 433--440.

\bibitem{fankhauser2018probabilistic}
P.~Fankhauser, M.~Bloesch, and M.~Hutter, ``Probabilistic terrain mapping for mobile robots with uncertain localization,'' \emph{IEEE Robotics and Automation Letters}, vol.~3, no.~4, pp. 3019--3026, 2018.

\bibitem{gangapurwala2021real}
S.~Gangapurwala, M.~Geisert, R.~Orsolino, M.~Fallon, and I.~Havoutis, ``Real-time trajectory adaptation for quadrupedal locomotion using deep reinforcement learning,'' in \emph{2021 IEEE International Conference on Robotics and Automation (ICRA)}.\hskip 1em plus 0.5em minus 0.4em\relax IEEE, 2021, pp. 5973--5979.

\bibitem{jain2020pixels}
D.~Jain, A.~Iscen, and K.~Caluwaerts, ``From pixels to legs: Hierarchical learning of quadruped locomotion,'' \emph{arXiv preprint arXiv:2011.11722}, 2020.

\bibitem{kareer2023vinl}
S.~Kareer, N.~Yokoyama, D.~Batra, S.~Ha, and J.~Truong, ``Vinl: Visual navigation and locomotion over obstacles,'' in \emph{2023 IEEE International Conference on Robotics and Automation (ICRA)}.\hskip 1em plus 0.5em minus 0.4em\relax IEEE, 2023, pp. 2018--2024.

\bibitem{schulman2017proximal}
J.~Schulman, F.~Wolski, P.~Dhariwal, A.~Radford, and O.~Klimov, ``Proximal policy optimization algorithms,'' \emph{arXiv preprint arXiv:1707.06347}, 2017.

\bibitem{rudin2022learning}
N.~Rudin, D.~Hoeller, P.~Reist, and M.~Hutter, ``Learning to walk in minutes using massively parallel deep reinforcement learning,'' in \emph{Conference on Robot Learning}.\hskip 1em plus 0.5em minus 0.4em\relax PMLR, 2022, pp. 91--100.

\bibitem{lee2020learning}
J.~Lee, J.~Hwangbo, L.~Wellhausen, V.~Koltun, and M.~Hutter, ``Learning quadrupedal locomotion over challenging terrain,'' \emph{Science robotics}, vol.~5, no.~47, p. eabc5986, 2020.

\bibitem{long2023hybrid}
J.~Long, Z.~Wang, Q.~Li, J.~Gao, L.~Cao, and J.~Pang, ``Hybrid internal model: A simple and efficient learner for agile legged locomotion,'' \emph{arXiv preprint arXiv:2312.11460}, 2023.

\bibitem{deeprobotics2024lite3}
``Deeprobotics lite3,'' \url{https://www.deeprobotics.cn/en/index/product1.html}, accessed on 2024-07-22.

\bibitem{deeprobotics2024joint}
``Deeprobotics j60 joint,'' \url{https://www.deeprobotics.cn/en/index/j60.html}, accessed on 2024-07-22.

\bibitem{unitree2024a1}
``Unitree a1,'' \url{https://www.unitree.com/a1}, accessed on 2024-07-22.

\bibitem{wang2023addressing}
W.~Wang, D.~Han, X.~Luo, and D.~Li, ``Addressing signal delay in deep reinforcement learning,'' in \emph{The Twelfth International Conference on Learning Representations}, 2023.

\end{thebibliography}

\end{document}